\documentclass{article}

\usepackage[final]{nips_2016}

\usepackage[utf8]{inputenc} % allow utf-8 input
\usepackage[T1]{fontenc}    % use 8-bit T1 fonts
\usepackage{url}            % simple URL typesetting
\usepackage{booktabs}       % professional-quality tables
\usepackage{amsfonts}       % blackboard math symbols
\usepackage{nicefrac}       % compact symbols for 1/2, etc.
\usepackage{microtype}      % microtypography
\usepackage{multirow}
\usepackage{rotating}
\usepackage{comment}
\usepackage{hyperref}       % hyperlinks

\title{Interpreting the Syntactic and Social Elements of the Tweet Representations via Elementary Property Prediction Tasks}

\author{
Ganesh J\\
  IIIT Hyderabad, India\\
  \texttt{ganesh.j@research.iiit.ac.in} \\
  \And
  Manish Gupta \\
  Microsoft, Hyderabad, India\\
  \texttt{gmanish@microsoft.com} \\
  \And
  Vasudeva Varma \\
  IIIT Hyderabad, India\\
  \texttt{vv@iiit.ac.in} \\
}

\begin{document}

\maketitle

\section{Introduction}
Research in social media analysis is recently seeing a surge in the number of research works applying representation learning models to solve high-level syntactico-semantic tasks such as sentiment analysis [1], semantic textual similarity computation [2], hashtag prediction [3] and so on. Though the performance of the representation learning models are better than the traditional models for all the tasks, little is known about the core properties of a tweet encoded within the representations. In a recent work, Hill et al. [4] perform a comparison of different sentence representation models by evaluating them for different high-level semantic tasks such as paraphrase identification, sentiment classification, question answering, document retrieval and so on. This type of coarse-grained analysis is opaque as it does not clearly reveal the kind of information encoded by the representations. Our work presented here constitutes the first step in opening the black-box of vector embeddings for social media posts, particularly tweets.

Essentially we ask the following question: ``What are the core properties encoded in the given tweet representation?''. We explicitly group the set of these properties into two categories: \textit{syntactic} and \textit{social}. Syntactic category includes properties such as tweet length, the order of words in it, the words themselves, slang words, hashtags and named entities in the tweet. On the other hand, social properties consist of `is reply', and `reply time'. We investigate the degree to which the tweet representations encode these properties. We assume that \textit{if we cannot train a classifier to predict a property based on its tweet representation, then this property is not encoded in this representation}. For example, the model which preserves the tweet length should perform well in predicting the length given the representation generated from the model. Though these elementary property prediction tasks are not directly related to any downstream application, knowing that the model is good at modeling a particular property (e.g., the social properties) indicates that it could excel in correlated applications (e.g., user profiling task). In this work we perform an extensive evaluation of 9 unsupervised and 4 supervised tweet representation models, using 8 different properties. The most relevant work is that of Adi et al. [5], which investigates three sentence properties in comparing unsupervised sentence representation models such as average of words vectors and LSTM auto-encoders. We differ from their work in two ways: (1) While they focus on sentences, we focus on social media posts which opens up the challenge of considering multiple salient properties such as hashtags, named entities, conversations and so on. (2) While they work with only unsupervised representation-learning models, we investigate the traditional unsupervised methods (BOW, LDA), unsupervised representation learning methods (Siamese CBOW, Tweet2Vec), as well as supervised methods (CNN, BLSTM).

Our main contributions are summarized below.
\begin{itemize}
\item Our work is the first towards interpreting the tweet embeddings in a fine-grained fashion. To this end, we propose a set of tweet-specific elementary property prediction tasks which help in unearthing the basic characteristics of different tweet representations.
\item To the best of our knowledge, this work is the first to do a holistic study of traditional, unsupervised and supervised representation learning models for tweets.
\item We compare various tweet representations with respect to such properties across various dimensions like tweet length and word ordering sensitivity.
\end{itemize}

The paper is organized as follows. Sections 2 and 3 discuss the set of proposed elementary property prediction tasks and the models considered for this study respectively. Section 4 and 5 presents the experiment setup and result analysis respectively. We conclude the work with a brief summary in Section 5.

\section{Elementary Property Prediction Tasks}
In this section we list down the set of proposed elementary property prediction tasks to test the characteristics of a tweet embedding. Table~\ref{tab:aux_task_list} explains all the tasks considered in this study. Note that we use a neural network to build the elementary property prediction task classifier which has the following two layers in order: the representation layer and the softmax layer on top whose size varies according to the specific task.  When there are more than one input for a task, we concatenate embeddings for each input.
%We use lower case italics (\textit{t}, \textit{ng}) to refer to tweet, ngram (with size denoted by `n') and boldface to refer to their corresponding vector representations (\textbf{t}, \textbf{ng}). We distinguish by indices when more than one element is considered ($\it{ng_1}, \it{ng_2}, \bf{ng_1}, \bf{ng_2}$). 

\begin{table}[h]
\centering
\scriptsize
\begin{tabular}{|p{0.05cm}|p{0.1cm}|p{1cm}|p{6cm}|p{6cm}|}
\hline
 & \textbf{Id} & \textbf{Name}  & \textbf{Objective}  & \textbf{Negative Sampling Method / Binning}  \\ \hline
\multirow{6}{*}{\begin{sideways}\textbf{Syntactic}\end{sideways}} & 1 & Length & Predict the number of words in the tweet & Use binned lengths with bin size of 4 \\ 
\cline{2-5}
\cline{2-5}
& 2 & Content & Predict whether the word is in the tweet or not & Randomly choose the word in the vocabulary but not in the tweet \\ \cline{2-5}
& 3 & Word Order & Predict whether the word $w_1$ appears before the word $w_2$ in the tweet or not & Flip the words $w_1$ and $w_2$   \\ \cline{2-5}
& 4 & Slang words \footnotemark[1] & Predict whether the n-gram $ng_2$ is the standardized form of the n-gram $ng_1$ in the tweet or not & Randomly sample the n-gram $ng_2$  \\ \cline{2-5}
& 5 & Hashtag & Predict whether the word is a hashtag in the tweet or not  & Randomly choose the word in the tweet which is not a hashtag  \\ \cline{2-5}
& 6 & NE [11] & Predict whether the n-gram is a Named Entity (NE) in the tweet or not & Randomly choose the n-gram in the tweet which is not a NE  \\ \hline
\multirow{2}{*}{\begin{sideways}\textbf{Social}\end{sideways}} & 7 & Is Reply & Predict whether the tweet is a reply tweet or not & Randomly choose a tweet that is a conversation starter \\ 
\cline{2-5}
\cline{2-5}
& 8 & Reply Time & Predict the number of minutes taken to get a reply for the tweet & Use binned minutes with bin size of 2 \\ \hline
\end{tabular}
\caption{Details of the Set of Proposed Elementary Property Prediction Tasks}
\label{tab:aux_task_list}
\end{table}

\footnotetext[1]{\url{https://noisy-text.github.io/norm-shared-task.html}}

\section{Representation Models}
\label{sec:rep_model}
In this section we list down the set of models considered in the study.
\subsection{Unsupervised}
\begin{itemize}
\item \textbf{Bag Of Words} (BOW) [17] - This simple representation captures the TF-IDF value of an n-gram. We pick top 50K n-grams, with the value of `n' going up to 5.
\item \textbf{Latent Dirichlet Allocation} (LDA) [18] - We use the topic distribution resulting by running LDA with number of topics as 200, as tweet representation.
\item \textbf{Bag Of Means} (BOM) - We take the average of the word embeddings obtained by running the Glove [12] model on 2 billion tweets with embedding size as 200.
\item \textbf{Deep Structured Semantic Models} (DSSM) [9] - This is a deep encoder trained to represent query and document in common space, for document ranking. We use the publicly available pre-trained encoder to encode the tweets.
\item \textbf{Convolutional DSSM} (CDSSM) [10] - This is the convolutional variant of DSSM.
\item \textbf{Paragraph2Vec} (PV) [13] - This model based on Word2Vec [15] learns embedding for a document which is good in predicting the words within it. We use the BOW variant with embedding size and window size of 200 and 10 respectively.
\item \textbf{Skip-Thought Vectors} (STV) [6] - This is a GRU [16] encoder trained to predict adjacent sentences in a books corpus. We use the recommended combine-skip (4800-dimensional) vectors from the publicly available encoder.
\item \textbf{Tweet2Vec} (T2V) [3] - This is a character composition model working directly on the character sequences to predict the user-annotated hashtags in a tweet. We use publicly available encoder, which was trained on 2 million tweets.
\item \textbf{Siamese CBOW} (SCBOW) [2] - This model uses averaging of word vectors to represent a sentence, and the objective and data used here is the same as that for STV. Note that this is different from BOW because the word vectors here are optimized for sentence representation.
\end{itemize}
\subsection{Supervised}
\begin{itemize}
\item \textbf{Convolutional Neural Network} (CNN) - This is a simple CNN proposed in [7].
\item \textbf{Long Short Term Memory Network} (LSTM) [14] - This is a vanilla LSTM based recurrent model, applied from start to the end of a tweet, and the last hidden vector is used as tweet representation.
\item \textbf{Bi-directional LSTM} (BLSTM) [14] - This extends LSTM by using two LSTM networks, processing a tweet left-to-right and right-to-left respectively. Tweet is represented by concatenating the last hidden vector of both the LSTMs.
\item \textbf{FastText} (FT) [8] - This is a simple architecture which averages the n-gram vectors to represent a tweet, followed by the softmax in the final layer. This simple model has been shown to be effective for the text classification task.
\end{itemize}

\section{Experiments}
\label{sec:exp}
In this section we perform an extensive evaluation of all the models in an attempt to find the significance of different representation models. Essentially we study every model (with optimal settings reported in the corresponding paper) with respect to the following three perspectives.
\begin{enumerate}
\item \textbf{Property Prediction Task Accuracy} - This test identifies the  model with the best F1-score for each elementary property prediction task. \begin{enumerate}
\item \textit{Best of all in}: Property prediction tasks for which this model has outperformed all the other models.
\item \textit{Best of unsupervised approaches in}: Property prediction tasks for which this model has outperformed all the other unsupervised models.
\item \textit{Best of supervised approaches in}: Property prediction tasks for which this model has outperformed all the other supervised models.
\end{enumerate}
\item \textbf{Property Prediction Task Accuracy versus Tweet Length} - This test helps to compare the performance of the model for shorter and longer tweets.
\begin{enumerate}
\item \textit{Positively correlated tasks}: Property prediction tasks for which the performance of the model increases as tweet length increases.
\item \textit{Negatively correlated tasks}: Property prediction tasks for which the performance of the model decreases as tweet length increases.
\end{enumerate}
\item \textbf{Sensitivity of Property Prediction Task to Word Order} - This refers to the setting where the words in the tweets are randomly ordered. This helps in testing the extent to which a model relies on the word ordering properties of the natural language.
\begin{enumerate}
\item \textit{Invariant tasks}: Property prediction tasks for which the model performance does not decline even when the words in the tweets are randomly reordered.
\item \textit{Significantly deviant tasks}: Property prediction tasks for which the model performance declines significantly when the words in the tweets are randomly reordered.
\end{enumerate}
\end{enumerate}

%\begin{comment}
\begin{table}[h]
\centering
\scriptsize
\begin{tabular}{|p{0.8cm}|p{3cm}|p{2.8cm}|p{3cm}|}
\hline
\textbf{Model} & \textbf{Task Accuracy (1)} & \textbf{Tweet Length (2)} & \textbf{Permuted (3)} \\ \hline
BOW & 
(a): Reply Time \newline
(b): Reply Time, Slang Words
&  
(a): Is Reply, Reply Time \newline
(b): Length, Word Order
&
BOW is invariant to word order  \\ \hline
LDA &
(a): Hashtag \newline
(b): Hashtag
&
(a): Is Reply, Reply Time
&
(a): Content, Hashtag, Reply time \newline
(b): Word Order  \\ \hline
BOM &
(a): Word Order, NE \newline
(b): Word Order, NE
&
(a): Is Reply, Reply Time \newline
(b): Length, Word Order
&
Word vector average is unaffected by the reordering of the words  \\ \hline
DSSM &
Best in None &
(a): Is Reply, Reply Time \newline
(b): Length, Word Order
&
(a): Length, Content \newline
(b): Is Reply \\ \hline
CDSSM &
(a): Hashtag \newline
(b): Hashtag &
(a): Is Reply, Reply Time \newline
(b): Length, Word Order
&
(a): Hashtag, Reply Time \newline
(b): Slang words  \\ \hline
PV &
Best in None &
(a): Is Reply, Reply Time \newline
(b): Length, Word Order
&
(b): Length \\ \hline
STV &
(a): Content, Is Reply \newline
(b): Content, Is Reply &
(a): Is Reply, Reply Time \newline
(b): Length, Word Order
&
(a): Slang Words  \\ \hline
T2V &
(b): Length &
(a): Is Reply, Reply Time \newline
(b): Length, Word Order
&
(b): Slang Words, Is Reply  \\ \hline
SCBOW &
(a): Hashtag \newline
(b): Hashtag &
(a): Is Reply, Reply Time \newline
(b): Length, Word Order
&
Word vector average is unaffected by the reordering of the words  \\ \hline
CNN & 
(c): Content, Word Order, Hashtag, Reply Time    &
(a): Is Reply, Reply Time \newline
(b): Length, Word Order
&
(a): Hashtag \newline
(b): Length, Slang Words, Is Reply, Reply Time
 \\ \hline
LSTM &
(a): Length  &
(a): Is Reply, Reply Time \newline
(b): Word Order
&
(a): NE \newline
(b): Slang Words, Is Reply
 \\ \hline
BLSTM &
(a): Slang Words \newline
(c): Is Reply, Slang Words &
(a): Is Reply, Reply Time \newline
(b): Word Order 
&
(b): Length, Is Reply, Reply Time
\\ \hline
FT &
Best in None &
(a): Is Reply, Reply Time \newline
(b): Length, Word Order
&
Word vector average is unaffected by the reordering of the words
 \\ \hline
\end{tabular}
\caption{Fine-grained analysis of supervised/unsupervised models}
\label{tab:analysis}
\end{table}
%\end{comment}

\section{Results and Analysis}
Fine-grained analysis of various supervised and unsupervised models discussed in Section~\ref{sec:rep_model}, across various dimensions discussed in Section~\ref{sec:exp}, is presented in Table~\ref{tab:analysis}. The codes used to conduct our experiments are publicly accessible at: \url{https://github.com/ganeshjawahar/fine-tweet/}.

\begin{table*}[]
\centering
\scriptsize
\begin{tabular}{|l|p{1.33cm}|p{0.7cm}|p{0.7cm}|p{0.7cm}|p{0.7cm}|p{0.95cm}|p{0.7cm}|p{0.8cm}|p{1.1cm}|}
\hline
 & \textbf{Model / Task} & \textbf{Len.} & \textbf{Cont.} & \textbf{Order} & \textbf{Slang} & \textbf{Hash tag} & \textbf{NE} & \textbf{Is Rep.} & \textbf{Rep. Time}  \\ \hline
\multirow{9}{*}{\begin{sideways}\textbf{Unsupervised}\end{sideways}} & 
BOW & 37.83 & 97.37 & 60.36 & 78.13 & 99.28 & 89.66 & 78.14 & \textbf{35.98}  \\  
\cline{2-10}
\cline{2-10}
& LDA & 25.11 & 97.72 & 60.62 & 76.82 & \textbf{99.35} & 97.24 & 60.12 & 28.03  \\ \cline{2-10}
& BOM & 47.64 & 98.67 & \textbf{61.25} & 75.26 & 99.33 & \textbf{98.06} & 66.25 & 28.43  \\ \cline{2-10}
& DSSM & 57.76 & 98.57 & 59.01 & 76.89 & 99.33 & 97.16 &  76.47 & 29.08 \\ \cline{2-10}
& CDSSM & 47.75 & 98.09 & 57.66 & 69.8 & \textbf{99.35} & 97.41 &  73.92 & 28.49  \\ \cline{2-10}
& PV & 13.58 & 94.9 & 60.92 & 76.09 & 85.61 & 98.02 & 54.68 & 27.58 \\ \cline{2-10}
& STV & 71.85 & \textbf{98.85} & 57.7 & 76.66 & 99.32 & 97.92 & \textbf{96.41} & 29.25  \\ \cline{2-10}
& T2V & 73.58 & 98.36 & 60.62 & 62.34 & 99.32 & 92.93 & 95.73 & 31.59 \\ \cline{2-10}
& SCBOW & 32.13 & 97.94 & 58.39 & 74.24 & \textbf{99.35} & 97.79 & 60.38 & 28.39  \\ \hline
\multirow{4}{*}{\begin{sideways}\textbf{Super.}\end{sideways}} & 
CNN & 59.48 & 97.71 & 61.13 & 77.42 & 99.31 & 91.38 & 92.66 & 31.73 \\ \cline{2-10}
& LSTM & \textbf{99.79} & 97.39 & 60.74 & 76.24 & 99.28 & 90.36 & 92.39 & 28.46 \\ \cline{2-10}
& BLSTM & 98.72 & 97.47 & 60.85 & \textbf{80.52} & 99.28 & 90.89 & 92.76 & 27.99\\ \cline{2-10}
& FT & 24.56 & 92.15 & 60.06 & 67.48 & 89.11 & 78.89 & 74.08 & 28.35 \\ \hline
\end{tabular}
\caption{Elementary Property Prediction Task F1-Score (\%) - Performance Comparison}
\label{tab:ta}
\end{table*}

\subsection{Property Prediction Task Accuracy}
We summarize the results of property prediction tasks in Table~\ref{tab:ta}. Length prediction turns out to be a difficult task for most of the models. Models which rely on the recurrent architectures such as LSTM, STV, T2V have sufficient capacity to perform well in modeling the tweet length. Also BLSTM is the best in modeling slang words. BLSTM outperforms the LSTM variant in all the tasks except `Content', which signifies the power of using the information flowing from both the directions of the tweet. T2V which is expected to perform well in this task because of its ability to work at a more fine level (i.e., characters) performs the worst. In fact T2V does not outperform other models in any task, which could be mainly due to the fact that the hashtags which are used for supervision in learning tweet representations reduces the generalization capability of the tweets beyond hashtag prediction. Prediction tasks such as `Content' and `Hashtag' seem to be less difficult as all the models perform nearly optimal for them. The superior performance of all the models for the `Content' task in particular is unlike the relatively lower performance reported for in [5], mainly because of the short length of the tweets. The most surprising result is when the BOM model turned out to be the best in `Word Order' task, as the model by nature loses the word order. This might be due to the correlation between word order patterns and the occurrences of specific words. BOM has also proven to perform well for identifying the named entities in the tweet.

STV is good for most of the social tasks. We believe the main reason for STV's performance is two-fold: (a) the inter-sentential features extracted from STV's encoder by the prediction of the surrounding sentences in the books corpus contains rich social elements that are vital for social tasks (e.g., user profiling), (b) the recurrent structure in both the encoder and decoder persists useful information in the memory nicely. The second claim is further substantiated by observing the poor performance of SCBOW whose objective is also similar to STV, but with a simpler architecture (i.e., word vector averaging). In future it would be interesting to create such a model for Twitter conversations or chronologically ordered topical tweets so as to directly capture the latent social features from Twitter.

\subsection{Property Prediction Task Accuracy versus Tweet Length}
This setup captures the behavior of the model with the increase in the context size, which is defined in terms of number of words. For `Word Order' task, we see the performance of all the models to be negatively correlated with the tweet length, which is expected. On the other hand, there is no correlation between the tweet length and the performance of all the models for the tasks such as `Slang Words', `Content', `Hashtag', `NE', and `Is Reply'. For social tasks such as `Is Reply' and `Reply Time', we see a positive correlation between the tweet length and the performance of all the models. This finding is intuitive in social media analysis where additional context is mostly helpful in modeling the social behavior.

\subsection{Sensitivity of Property Prediction Task to Word Order}
This test essentially captures the importance of ``natural word order''. We found that LDA was invariant to the reordering of the words in the tweet for most of the tasks. This result is not surprising as LDA considers each word in the tweet independently. CNN, LSTM and BLSTM rely on the word order significantly to perform well for most of the prediction tasks.

\begin{comment}
\begin{itemize}
\item For content prediction task, all the models nearly performs equally and competitively. This result deviates from general sentences [5], and is mainly due to the short lengths of the tweets.
\item PV and DSSM models do not excel other models in any task.
\item Performance for word order task is negatively correlated with respect to tweet length for all the models.
\item Performance for tasks such as content, hashtag, NE and `is reply' have no correlation with respect to tweet length for all the models.
\item For identifying the slang words, we discover that there is no correlation between the tweet length and performance of any model.
\item Performance for both the social tasks are positively correlated with respect to tweet length for all the models. This phenomena explains why more context is always helpful for modeling the social behavior.
\item LDA has proven to be invariant to the reordering of the words in the tweet for most of the tasks. This result is not surprising as LDA considers each word in the tweet independently.
\item CNN, LSTM and BLSTM rely on the properties of the natural language significantly to perform well for most of the elementary tasks.
\end{itemize}
\end{comment}

\section{Conclusion}
This work proposed a set of elementary property prediction tasks to understand different tweet representations in an application independent, fine-grained fashion. The open nature of social media not only poses a plethora of opportunities to understand the basic characteristics of the posts, but also helped us draw novel insights about different representation models. We observed that among supervised models, CNN, LSTM and BLSTM encapsulates most of the syntactic and social properties with a great accuracy, while BOW, DSSM, STV and T2V does that among the unsupervised models. Tweet length affects the task prediction accuracies, but we found that all models behave similarly under variation in tweet length. Finally while LDA is insensitive to input word order, CNN, LSTM and BLSTM are extremely sensitive to word order.
%In future we wish to generalize the approach to more complex syntactic and social properties.

\section*{References}
%\medskip
%\small

[1] Tang, D., Wei, F., Qin, B., Yang, N., Liu, T., \& Zhou, M.: Sentiment Embeddings with Applications to Sentiment Analysis. In: TKDE. (2016) 28(2) 496-509

[2] Kenter, T., Borisov, A., \& de Rijke, M.: Siamese CBOW: Optimizing Word Embeddings for Sentence Representations. In: ACL. (2016) 941-951

[3] Dhingra, B., Zhou, Z., Fitzpatrick, D., Muehl, M., \& Cohen, W. W.: Tweet2Vec: Character-Based Distributed Representations for Social Media. In: ACL. (2016)

[4] Hill, F., Cho, K.,\& Korhonen, A.: Learning distributed representations of sentences from unlabelled data. In: NAACL. (2016)

[5] Adi, Y., Kermany, E., Belinkov, Y., Lavi, O., \& Goldberg, Y.: Fine-grained Analysis of Sentence Embeddings Using Auxiliary Prediction Tasks. arXiv preprint arXiv:1608.04207. (2016)

[6] Kiros, R., Zhu, Y., Salakhutdinov, R. R., Zemel, R., Urtasun, R., Torralba, A., \& Fidler, S.: Skip-thought vectors. In: NIPS. (2015) 3294-3302

[7] Kim, Y.: Convolutional neural networks for sentence classification. In: EMNLP. (2014)

[8] Joulin, A., Grave, E., Bojanowski, P., \& Mikolov, T.: Bag of Tricks for Efficient Text Classification. arXiv preprint arXiv:1607.01759. (2016)

[9] Huang, P. S., He, X., Gao, J., Deng, L., Acero, A., \& Heck, L.: Learning deep structured semantic models for web search using clickthrough data. In: CIKM. (2013)

[10] Shen, Y., He, X., Gao, J., Deng, L., \& Mesnil, G.: A latent semantic model with convolutional-pooling structure for information retrieval. In: CIKM. (2014)

[11] Ritter, A., Clark, S., Mausam, \& Etzioni, O.: Named entity recognition in tweets: an experimental study. In: EMNLP. (2011) 1524-1534

[12] Pennington, J., Socher, R., \& Manning, C. D.: Glove: Global Vectors for Word Representation. In: EMNLP. (2014) 1532-43

[13] Le, Q. V., \& Mikolov, T.: Distributed Representations of Sentences and Documents. In: ICML. (2014) 1188-1196

[14] Graves, A., Mohamed, A. R., \& Hinton, G.: Speech recognition with deep recurrent neural networks. In: ICASSP. (2013) 6645-6649

[15] Mikolov, T., Sutskever, I., Chen, K., Corrado, G. S., \& Dean, J.: Distributed representations of words and phrases and their compositionality. In: NIPS. (2013) 3111-3119

[16] Cho, K., Van Merriënboer, B., Bahdanau, D., \& Bengio, Y.: On the properties of neural machine translation: Encoder-decoder approaches. arXiv preprint arXiv:1409.1259. (2014)

[17] Harris, Z. S.: Distributional structure. In: Word. (1954) 146-162

[18] Blei, D. M., Ng, A. Y., \& Jordan, M. I.: Latent dirichlet allocation. In: JMLR. (2003)

\end{document}